\documentclass[letterpaper]{article} 
\usepackage[preprint]{aaai2027} 

\usepackage[hyphens]{url}
\usepackage{graphicx}
\usepackage{natbib} 
\usepackage{caption} 
\usepackage{amsmath}
\usepackage{amssymb}
\usepackage{amsfonts}

\usepackage{booktabs}
\usepackage{multirow}
\usepackage{threeparttable}
\usepackage{tablefootnote}
\usepackage{colortbl}

\usepackage{enumitem}

\usepackage{algorithm}
\usepackage{algpseudocode}

\usepackage{framed}
\usepackage{mdframed}
\usepackage[most]{tcolorbox}
\usepackage{listings}
\usepackage{inconsolata}

\usepackage{xcolor}
\usepackage{pifont}

\definecolor{myGreen}{RGB}{34,139,34}
\definecolor{myRed}{RGB}{220,20,60}

\definecolor{cvrpRed1}{RGB}{253,231,218}
\definecolor{cvrpRed2}{RGB}{252,207,182}
\definecolor{cvrpRed3}{RGB}{251,177,146}

\definecolor{cvrpGreen1}{RGB}{198,232,192}
\definecolor{cvrpGreen2}{RGB}{165,219,155}

\definecolor{cvrpBlue1}{RGB}{70,130,180}
\definecolor{cvrpBlue2}{RGB}{40,90,160}

\title{Self-Guided LLM Reasoning for Error Diagnosis in Vehicle Routing Problems}

\author{
Ni Zhang\textsuperscript{\rm 1},
Jianan Zhou\textsuperscript{\rm 2},
Yue-Jiao Gong\textsuperscript{\rm 3},
Zhiguang Cao\textsuperscript{\rm 1}
}

\affiliations{
\textsuperscript{\rm 1}Singapore Management University\quad
\textsuperscript{\rm 2}Nanyang Technological University\quad
\textsuperscript{\rm 3}South China University of Technology\\
\vspace{2mm}
ni.zhang.2025@phdcs.smu.edu.sg,\quad
jianan004@e.ntu.edu.sg,\\
gongyuejiao@gmail.com,\quad
zgcao@smu.edu.sg
}

\begin{document}
\maketitle
\begin{abstract}
Vehicle Routing Problems (VRPs) are fundamental to many real-world applications. Recent advances in large language models (LLMs) have spurred progress in solving these NP-hard problems. However, most existing approaches assume that input instances are error-free, neglecting the presence of faults commonly encountered in real-world data. In this work, we systematically investigate the capability of LLMs to detect (and correct) errors in VRP instances, and identify key limitations, particularly in handling complex instances or errors that substantially increase the LLM's reasoning burden. To overcome these challenges, we propose a self-guided reasoning framework in which LLMs autonomously generate code that provides structured insights to aid in error diagnosis. Experiments across diverse problem settings and LLMs demonstrate that our framework significantly improves error detection performance, achieving near-perfect accuracy.
\end{abstract}

\section{Introduction}
Vehicle Routing Problems (VRPs) are widely deployed in real-world decision-making pipelines across last-mile delivery~\citep{tiwari2023optimization,tilk2021last,kitjacharoenchai2019vehicle}, transportation~\citep{gayialis2019vehicle,wang2024dual,xu2023manipulating}, and ride-sharing~\citep{lin2012research,guo2022vehicle,chen2019ride}. While decades of research have produced highly efficient optimization algorithms, the reliability of these pipelines remains fragile: subtle inconsistencies or latent errors in problem instances can lead to solver failures or infeasible solutions.
Most VRP solvers assume that input instances are well-formed and feasible. In practice, however, instances may contain errors such as formatting issues, missing fields, or violated feasibility constraints. Importantly, many of these faults are \emph{latent} as they are not immediately evident from the instance specification and require additional analysis to be identified.
Given the diversity of VRP formulations and input formats across problem settings, designing general-purpose programs for automatic error detection is highly challenging.
As a result, prior approaches typically rely on domain expertise or post hoc analysis of solver outputs to infer potential instance-level issues, which is both time-consuming and expertise-dependent.

Recent advances in large language models (LLMs) offer a promising new direction. With strong language understanding and reasoning capabilities, LLMs can interpret diverse inputs and reason about structured information. These capabilities naturally prompt following questions: \emph{a) Can LLMs be enabled to identify and correct errors in VRP instances? b) How and why does such reasoning succeed or fail? c) Under what conditions do LLMs struggle to identify errors, and how can these limitations be effectively mitigated?}

Recent work has begun to explore this direction. For example, MOID~\citep{li2025multi} attempts to identify potential instance-level issues by examining the feasibility of solutions produced by a solver. However, it heavily relies on solver outputs and thus inherits their limitations, since infeasible solutions may arise from solver deficiencies rather than errors in the input instance. Moreover, a comprehensive evaluation of LLMs' capability to directly identify and correct errors within VRP instances remains absent to date.

In this paper, we systematically investigate the capability of LLMs to diagnose errors in VRP instances. To address the aforementioned research questions, we evaluate: (i) LLMs' ability to handle increased error complexity, (ii) their ability to cope with increased instance complexity, and (iii) the effect of iterative reasoning on error detection performance.
Our findings show that (i) LLMs perform well in detecting formatting errors and relatively simple constraint violations, such as capacity and demand constraints. However, their ability to identify more complex constraint violations, including time windows and distance limits, as well as scenarios with larger numbers of errors, is substantially weaker. (ii) Detection performance degrades as instance scale and constraint counts increases, indicating limited robustness under increased instance complexity. (iii) While iterative reasoning improves the detection of formatting issues and simple constraint violations, it provides limited benefit for identifying more complex errors.

In summary, we find that LLMs are effective at identifying and correcting errors in relatively simple settings, whereas their performance degrades when confronted with instances or errors of higher complexity. To address limitations, we propose a self-guided reasoning framework in which the LLM generates analysis code as an intermediate reasoning tool. This code enables the model to extract structured insights that facilitate error detection while alleviating overall reasoning burden. Our contributions are summarized as follows:
\begin{itemize}
    \item We present a comprehensive study of LLMs' capability to detect errors in VRP instances, including a systematic analysis of how error complexity, instance complexity and reasoning depth affect detection performance.
    \item We propose a self-guided reasoning framework in which LLMs generate code to derive intermediate reasoning insights, addressing their limitations in handling complex errors while reducing the overall reasoning burden.
    \item Extensive experiments across diverse problem settings, including vehicle routing and bin packing, demonstrate that our framework significantly enhances error detection performance, achieving near-perfect accuracy.
\end{itemize}

\section{Related work}
Existing uses of LLMs in VRPs broadly fall into two categories. The first line of work focuses on using LLMs to directly generate solutions. For example, OPRO~\citep{yang2024large} employs LLMs to produce solutions directly, LEMA~\citep{liu2024large} leverages LLMs to perform genetic search, and Jiang et al.~\citep{jiang2025large} fine-tune LLMs for solution generation. The second line of work instead leverages LLMs to generate heuristic code or optimization operators for solving VRPs. FunSearch~\citep{romera2024mathematical} adopts an island-based evolutionary framework to guide LLMs in generating heuristic programs, while EOH~\citep{liu2024evolution} integrates LLMs with evolutionary algorithms to automatically discover improved heuristics. ReEvo~\citep{ye2024reevo} further augments the evolutionary loop with reflective self-feedback, enabling the LLM to critique and refine its own proposals, and HSEvo~\citep{dat2025hsevo} introduces a hierarchical evolutionary structure with diversity-aware selection over modular heuristic components. AFL~\citep{zhang2025agentic} extends this paradigm with an agentic framework for solving complex VRPs through multi-stage reasoning and coding. Overall, these works demonstrate the growing potential of LLMs.

However, all of these methods assume that input instances are feasible and error-free, and therefore lack the capability to handle instance-level errors that frequently arise in practice. MOID~\citep{li2025multi} takes a step toward error detection by analyzing solver execution, but its reliance on potentially unreliable solvers makes it difficult to distinguish between instance-level errors and solver-induced failures. As such, it serves as an initial attempt rather than a comprehensive solution. In this work, we bridge this gap by enabling LLMs to directly identify and reason about instance-level errors.

\section{A Comprehensive Study}
In this section, we conduct a comprehensive study on LLMs' capability to detect error in VRP instances, examining performance across varying levels of error complexity, instance complexity, and reasoning depth. To enable a systematic evaluation, we construct error-injected VRP datasets by introducing 200 errors into 20 instances for each VRP variant. These errors include both formatting errors (e.g., value errors or missing entries) and feasibility-related errors (e.g., constraint violation). Detailed descriptions are provided in Appendix~\ref{sec:dataset}. 

\begin{figure}[t]
    \centering

    \includegraphics[width=1.0\linewidth]{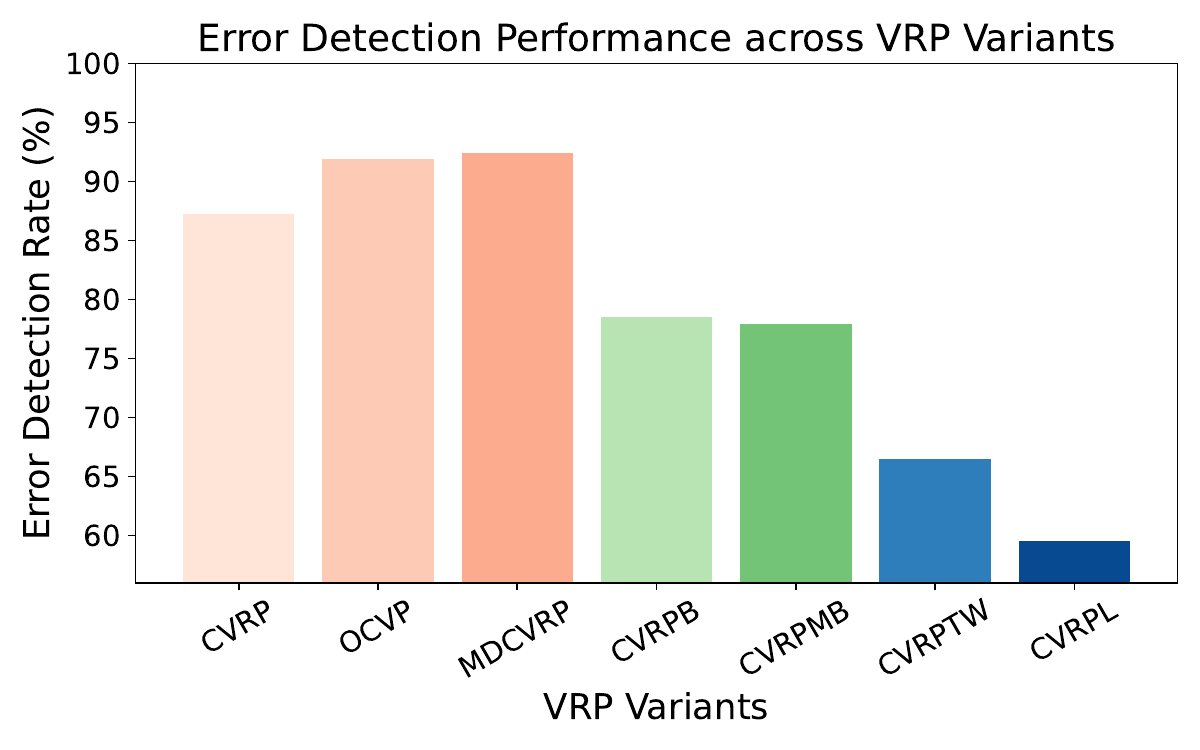}

    \vspace{0.8em}

    \resizebox{\linewidth}{!}{
    \begin{tabular}{l|ccccc}
    \toprule
    Variant & Format & Capacity & Backhaul & Distance Limit & Time Window \\
    \midrule
    CVRP    & \cellcolor{cvrpRed1}91.38 & \cellcolor{cvrpRed1}84.28 & --    & --    & -- \\
    OCVRP   & \cellcolor{cvrpRed2}90.32 & \cellcolor{cvrpRed2}92.38 & --    & --    & -- \\
    MDCVRP  & \cellcolor{cvrpRed3}94.44 & \cellcolor{cvrpRed3}87.67 & --    & --    & -- \\
    CVRPB   & \cellcolor{cvrpGreen1}98.90 & --    & \cellcolor{cvrpGreen1}61.19 & --    & -- \\
    CVRPMB  & \cellcolor{cvrpGreen2}91.18 & --    & \cellcolor{cvrpGreen2}64.29 & --    & -- \\
    CVRPTW  & \cellcolor{cvrpBlue1}95.38 & \cellcolor{cvrpBlue1}90.00 & --    & --    & \cellcolor{cvrpBlue1}0.00 \\
    CVRPL   & \cellcolor{cvrpBlue2}97.53 & \cellcolor{cvrpBlue2}82.97 & --    & \cellcolor{cvrpBlue2}0.00 & -- \\
    \bottomrule
    \end{tabular}
    }

    \caption{Overall error detection performance across different VRP variants, with a breakdown by error type shown below (higher is better).}
    \label{fig:vrp_fault_detection}
\end{figure}

\subsection{Error Complexity}
We analyze LLMs' ability to handle varying levels of error complexity, including constraint violations across different problem variants and varying numbers of errors.

\textbf{Problem Variant.} We analyze error detection performance across 7 simple VRP variants, each containing one or two types of constraints. 
As shown in Fig.~\ref{fig:vrp_fault_detection}, LLMs achieve consistently high detection rates for formatting errors and faults associated with simple constraints (e.g., capacity (C)), indicated by the red bars. These results suggest that LLMs are effective at identifying errors that can be detected through basic inspection or straightforward logical checks.

However, detection rates degrade substantially as constraint complexity increases. For variants involving backhaul (B) constraints, accuracy declines noticeably. For more complex constraints, such as time windows (TW) and distance limits (L), detection rates drop sharply and approach zero. 
This highlights a key limitation: while LLMs perform well on formatting errors and simple constraint violations, they struggle with errors that require complex calculations or numerical reasoning.

\begin{figure}[t]
    \centering
    \includegraphics[width=0.9\linewidth]{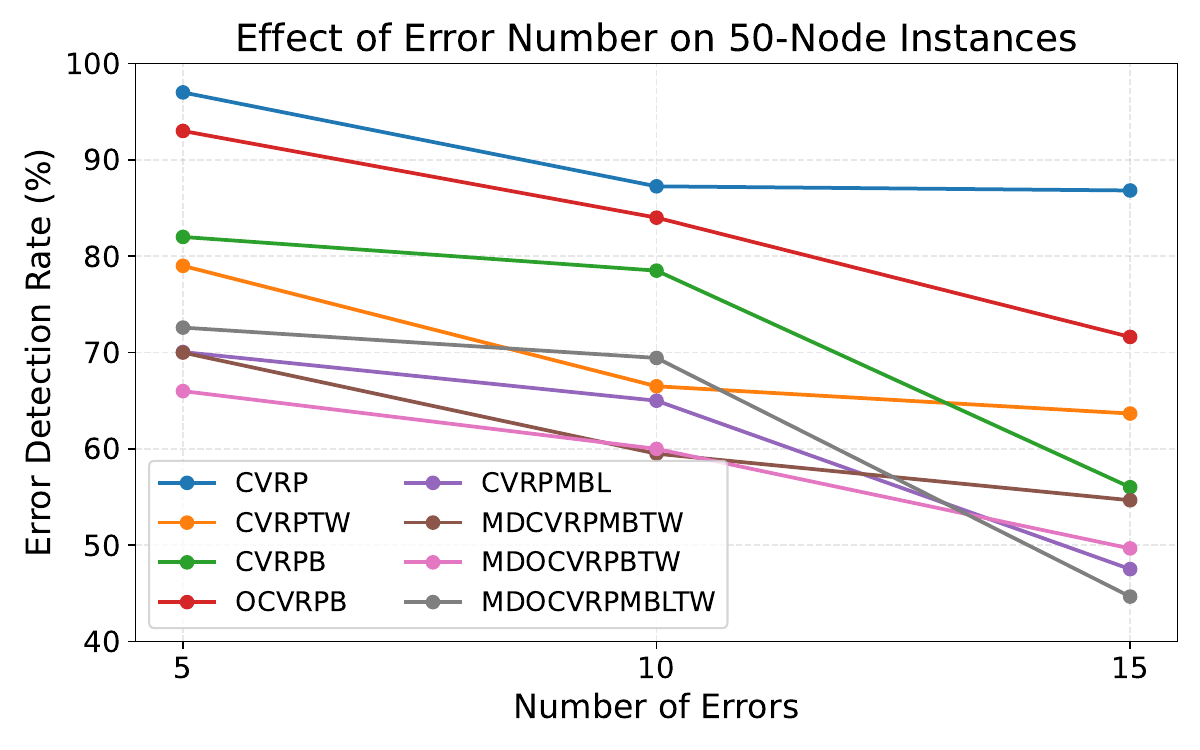}
    \includegraphics[width=0.9\linewidth]{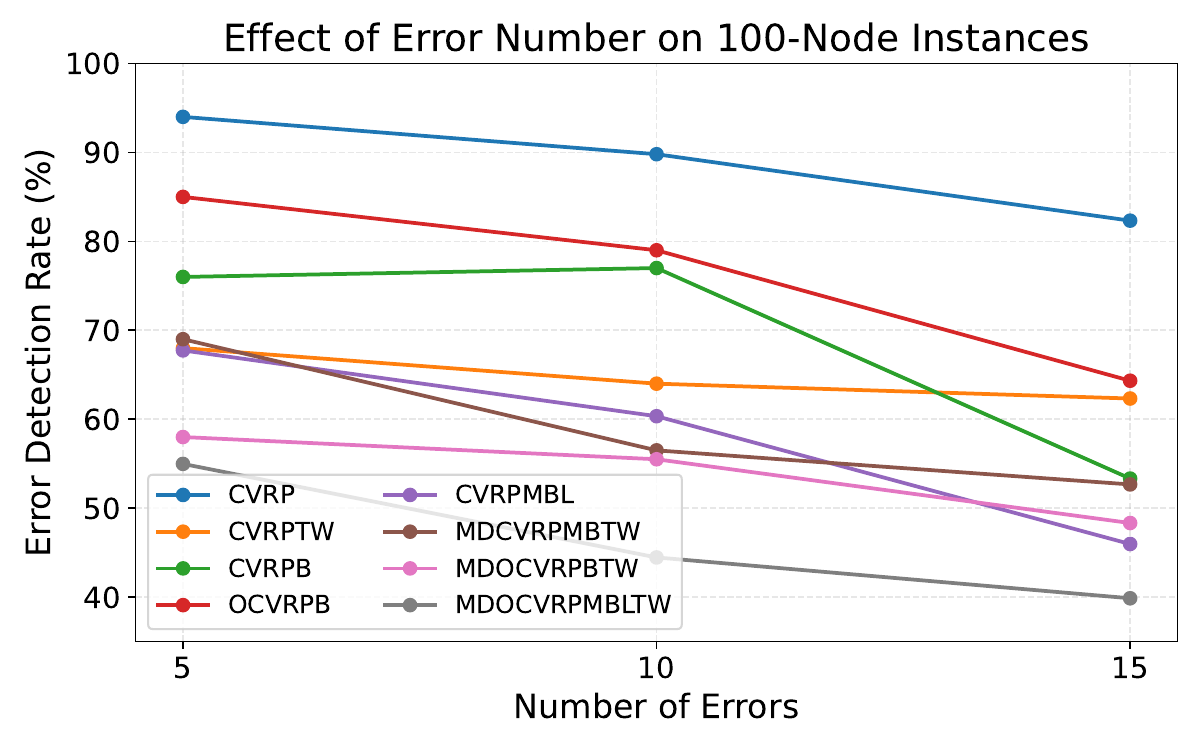}
    \caption{Effect of fault number and instance scale.}
    \label{fig:fault_quantity}
\end{figure}

\textbf{Error Number.} We further evaluate the impact of error number on LLM-based error detection by varying the number of injected errors and measuring detection performance across different VRP variants. To provide a more comprehensive assessment of its impact, we select variants with increasing constraint complexity, ranging from CVRP, which involves a single type of constraint, to MDOCVRPMBLTW, which incorporates six types of constraints.

Fig.~\ref{fig:fault_quantity} shows that detection performance consistently declines as the number of injected errors increases across all variants. This trend holds regardless of the underlying problem type, indicating that LLMs are highly sensitive to the accumulation of errors within an instance. These results suggest that a larger number of errors imposes a substantially greater reasoning burden on LLMs, as they must simultaneously identify more errors.

\subsection{Instance Complexity}
Beyond error complexity, intrinsic instance complexity may also impact LLMs' ability to detect errors. To study this, we conduct experiments across varying instance scales and constraint counts.

\begin{figure}[t]
    \centering
    \includegraphics[width=1.0\linewidth]{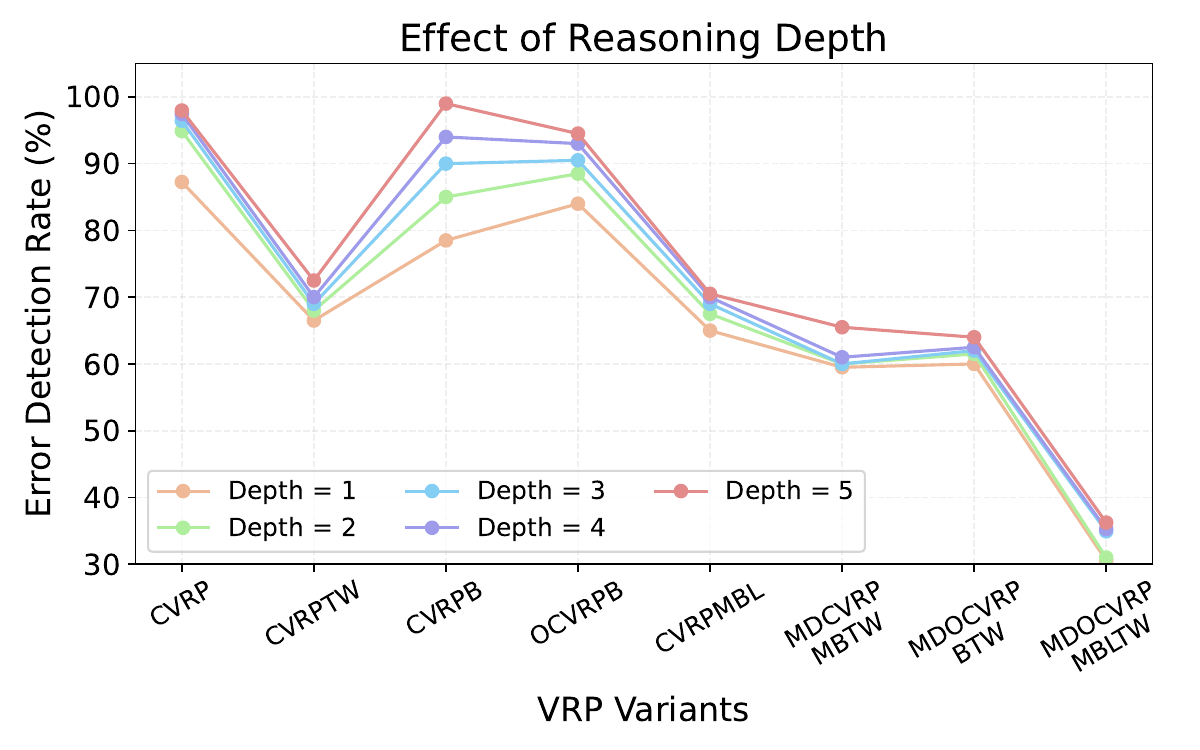}
    \caption{Effect of reasoning depth on error detection performance across different VRP variants.}
    \label{fig:reasoning_depth}
\end{figure}

\textbf{Instance Scale.} We examine the effect of instance scale on error detection performance by comparing instances of different sizes. As shown in Fig.~\ref{fig:fault_quantity}, detection rates on 100-node instances are uniformly lower than those on 50-node instances across all VRP variants and error settings.
This performance degradation indicates that larger instance significantly increase the complexity of instance-level reasoning. As the number of nodes grows, LLMs reason over a larger set of constraints and dependency relationships, which exacerbates the reasoning burden and makes accurate error identification increasingly difficult.

\begin{figure*}[t]
    \centering
    \includegraphics[width=\textwidth,trim=0cm 9.5cm 7cm 0cm, clip]{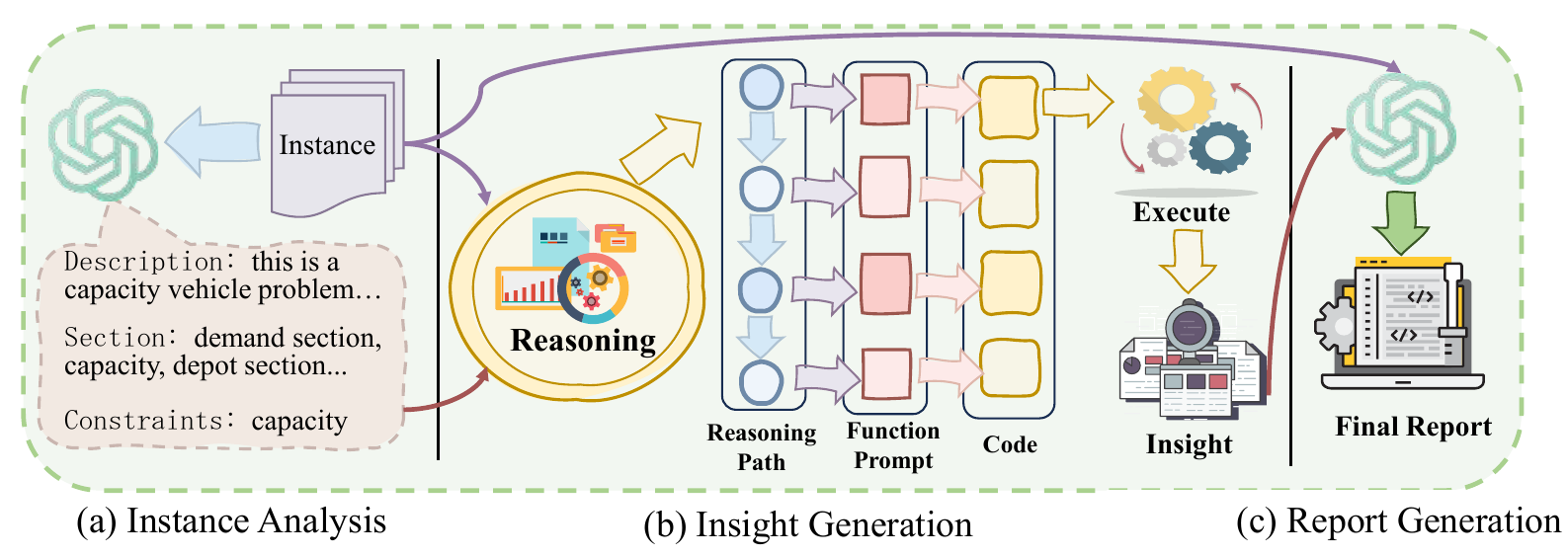}
    \caption{Overall framework of Self-Guided Reasoning. }
    \label{fig:framework}
\end{figure*}
\textbf{Count of Constraint Type.} We evaluate LLM-based error detection across 48 VRP variants with constraint counts ranging from one to six, and report the average detection rate for each constraint level. As shown in Table~\ref{tab:constraint_count}, error detection performance exhibits a clear decreasing trend as the number of constraints types increases. While LLMs achieve relatively high detection accuracy when only one or two constraint types are present, performance degrades substantially as more constraint types are introduced, with a sharp drop observed for instances involving five or six constraint types. These results indicate that an increasing number of constraint types significantly challenges LLMs' ability to identify errors, highlighting a key limitation of direct LLM-based error detection in highly constrained VRPs.
\begin{table}[t]
\centering
\caption{Average error detection rates of LLMs under different numbers of constraint types. (higher is better).}
\label{tab:constraint_count}
\resizebox{\linewidth}{!}{
\setlength{\tabcolsep}{8pt}
\renewcommand{\arraystretch}{1.3}
\begin{tabular}{c|cccccc}
\hline
 Constraint Type 
& 1 & 2 & 3 & 4 & 5 & 6 \\
\hline
LLM (\%) 
& \cellcolor{orange!70}87.27 
& \cellcolor{orange!45}79.42 
& \cellcolor{orange!35}72.42 
& \cellcolor{orange!20}57.42 
& \cellcolor{orange!15}53.97 
& \cellcolor{orange!10}22.30 \\
\hline
\end{tabular}
}
\end{table}

\subsection{Reasoning Depth}
Iterative reasoning is an effective approach to improve error detection. To evaluate its impact, we analyze the role of reasoning depth through an iterative process, in which the LLM identifies errors, the instance is repaired accordingly, and the LLM then reasons again on the updated version. As shown in Fig.~\ref{fig:reasoning_depth}, increasing reasoning depth consistently improves detection rates across VRP variants. For variants with relatively simple constraints (e.g., CVRP and CVRPB), repeated reasoning yields performance gains, with detection rates approaching 100\% after several iterations.

However, for problems with more complex constraints, the benefits of deeper reasoning diminish. In particular, for instances with constraints such as time windows and combinations of multiple constraints, detection performance remains far from perfect. These results suggest that while iterative reasoning can facilitate error detection by LLMs, it remains insufficient for handling complex errors, such as violations involving intricate constraints or combinations of constraints that require sophisticated feasibility reasoning.

\section{Self-Guided Reasoning Framework}
\label{sec:Methodology}
Our analysis of LLMs' ability to detect instance-level errors reveals a key limitation: their difficulty in handling instances or errors of higher complexity which may require advanced reasoning.
To address this, we propose a self-guided reasoning framework in which the LLM first constructs a reasoning chain for error detection and then uses it to guide the generation of analysis code. The generated code yields structured insights that the LLM can reference, enhancing its ability to identify errors while alleviating the overall reasoning burden.

As illustrated in Fig.~\ref{fig:framework}, the framework begins with instance analysis, where the LLM analyzes the input instance by identifying the problem type, required section names, and associated constraints. Based on this analysis, the LLM derives a reasoning path that outlines how to analyze potential errors in the instance and generates corresponding function-level prompts for each step. Guided by these prompts, executable code is generated and run to extract relevant insights. Finally, using both the original instance and the extracted insights as references, the LLM produces a final report.

\subsection{Instance Analysis}
Given an input instance, the instance analysis stage aims to establish a concise and structured understanding of the problem specification. 
The LLM first parses the raw instance to identify its essential characteristics, which are summarized as follows:
\begin{itemize}
    \item \textbf{Problem type}, indicating the underlying optimization problem (e.g., CVRP) implied by the instance.
    \item \textbf{Required section names}, specifying the mandatory sections that a valid instance of the identified problem type should contain.
    \item \textbf{Associated constraints}, describing the feasibility conditions in the identified problem, such as capacity, time window and so on.
\end{itemize}

Based on this structured information, the LLM performs a self-validation step to identify potential inconsistencies among the problem type, required sections, associated constraints, and the instance content, and revises its analysis accordingly. 
Finally, the LLM produces an instance analysis summary, providing an intuitive and reliable reference that helps reduce the reasoning burden in subsequent stages. The detailed prompt used for instance analysis is provided in Appendix~\ref{sec:prompt_Instance Analysis}.

\begin{table*}[ht]
\centering
\caption{Error Detection Rate (EDR) comparison across 48 VRP variants (higher is better).}
\vspace{2mm}
\renewcommand\arraystretch{1.2}
\resizebox{1.0\linewidth}{!}{
\begin{threeparttable}
\begin{tabular}{l|cc|cc|cc|cc|cc|cc}
\toprule
& EDR(\%) & Time(m) & EDR(\%) & Time(m) & EDR(\%) & Time(m) & EDR(\%) & Time(m) & EDR(\%) & Time(m) & EDR(\%) & Time(m)\\
\midrule
& \multicolumn{2}{c|}{CVRP} & \multicolumn{2}{c|}{OCVRP} & \multicolumn{2}{c|}{CVRPTW} & \multicolumn{2}{c|}{CVRPL} & \multicolumn{2}{c|}{MDCVRP} & \multicolumn{2}{c}{CVRPB}\\
LLM & 87.25 & 6.35 & 91.88& 6.20 & 66.50 & 7.00 & 59.50 & 7.38 & 94.42 & 8.15 & 78.50 & 8.25 \\
Self-Guided & \cellcolor{orange!20}99.49 & 6.39 & \cellcolor{orange!20}97.98 & 7.20 & \cellcolor{orange!20}95.00 & 7.22 & \cellcolor{orange!20}100.00 & 8.56 & \cellcolor{orange!20}97.47 & 9.54 & \cellcolor{orange!20}94.00 & 9.03\\
\midrule
& \multicolumn{2}{c|}{CVRPMB} & \multicolumn{2}{c|}{OCVRPTW}
& \multicolumn{2}{c|}{MDOCVRP}
& \multicolumn{2}{c|}{OCVRPB}
& \multicolumn{2}{c|}{CVRPLTW}
& \multicolumn{2}{c}{MDCVRPL} \\
LLM & 77.95 & 9.45 & 79.50 & 10.96
& 94.50 & 8.17
& 84.00 & 8.55
& 59.61 & 9.21
& 59.11 & 8.50 \\
Self-Guided & \cellcolor{orange!20}96.92 & 9.97& 98.00 & 10.87
& \cellcolor{orange!20}98.50 & 8.32
& \cellcolor{orange!20}94.00 & 8.52
& \cellcolor{orange!20}99.51 & 8.93
& \cellcolor{orange!20}94.09 & 9.45 \\
\midrule
& \multicolumn{2}{c|}{CVRPBL}
& \multicolumn{2}{c|}{CVRPBTW}
& \multicolumn{2}{c|}{MDCVRPTW}
& \multicolumn{2}{c|}{MDCVRPB}
& \multicolumn{2}{c|}{OCVRPMB}
& \multicolumn{2}{c}{CVRPMBL} \\

LLM
& 56.50 & 7.04
& 64.00 & 13.96
& 76.00 & 15.48
& 71.50 & 10.57
& 86.00 & 9.27
& 65.00 & 7.63 \\

Self-Guided
& \cellcolor{orange!20}100.00 & 8.10
& \cellcolor{orange!20}95.00 & 12.36
& \cellcolor{orange!20}95.50 & 17.29
& \cellcolor{orange!20}95.50 & 11.35
& \cellcolor{orange!20}95.00 & 10.55
& \cellcolor{orange!20}91.50 & 7.93 \\
\midrule
& \multicolumn{2}{c|}{OCVRPL}
& \multicolumn{2}{c|}{CVRPMBTW}
& \multicolumn{2}{c|}{MDCVRPMB}
& \multicolumn{2}{c|}{MDCVRPMBL}
& \multicolumn{2}{c|}{CVRPMBLTW}
& \multicolumn{2}{c}{MDCVRPBTW}\\
LLM
& 75.25 & 7.90
& 64.00 & 10.46
& 78.00 & 8.64
& 44.50 & 9.55
& 61.39 & 12.43
& 51.00 & 12.50\\
Self-Guided
& \cellcolor{orange!20}99.50 & 7.38
& \cellcolor{orange!20}96.50 & 11.95
& \cellcolor{orange!20}96.00 & 8.88
& \cellcolor{orange!20}96.65 & 11.22
& \cellcolor{orange!20}98.51 & 12.17
& \cellcolor{orange!20}90.50 & 13.08\\
\midrule
& \multicolumn{2}{c|}{OCVRPLTW}
& \multicolumn{2}{c|}{MDOCVRPL}
& \multicolumn{2}{c|}{OCVRPBL}
& \multicolumn{2}{c|}{OCVRPMBL}
& \multicolumn{2}{c|}{MDOCVRPTW}
& \multicolumn{2}{c}{OCVRPBTW} \\

LLM
& 56.59 & 10.87
& 43.69 & 7.68
& 62.00 & 7.95
& 63.00 & 8.63
& 68.50 & 13.25
& 71.50 & 9.60 \\

Self-Guided
& \cellcolor{orange!20}96.10 & 12.95
& \cellcolor{orange!20}93.00 & 8.97
& \cellcolor{orange!20}96.50 & 8.56
& \cellcolor{orange!20}94.50 & 9.56
& \cellcolor{orange!20}98.50 & 14.12
& \cellcolor{orange!20}94.00 & 11.20 \\
\midrule
& \multicolumn{2}{c|}{OCVRPMBTW}
& \multicolumn{2}{c|}{MDOCVRPB}
& \multicolumn{2}{c|}{MDOCVRPMB}
& \multicolumn{2}{c|}{MDCVRPLTW}
& \multicolumn{2}{c|}{MDCVRPBL}
& \multicolumn{2}{c}{CVRPBLTW} \\

LLM
& 60.00 & 11.83
& 86.50 & 10.28
& 67.50 & 9.66
& 23.08 & 11.32
& 44.00 & 8.37
& 56.00 & 13.32 \\

Self-Guided
& \cellcolor{orange!20}99.50 & 13.35
& \cellcolor{orange!20}92.00 & 9.73
& \cellcolor{orange!20}97.00 & 8.80
& \cellcolor{orange!20}98.56 & 12.38
& \cellcolor{orange!20}90.00 & 10.62
& \cellcolor{orange!20}96.59 & 13.80 \\
\midrule
& \multicolumn{2}{c|}{MDVRPMBTW}
& \multicolumn{2}{c|}{MDOVRPLTW}
& \multicolumn{2}{c|}{OCVRPBLTW}
& \multicolumn{2}{c|}{OCVRPMBLTW}
& \multicolumn{2}{c|}{MDOCVRPBTW}
& \multicolumn{2}{c}{MDOCVRPMBTW} \\
LLM& 59.50 & 14.42
& 43.75 & 8.50
& 56.96 & 10.10
& 66.07 & 10.28
& 60.00 & 10.53
& 62.50 & 12.64 \\
Self-Guided & \cellcolor{orange!20}90.00 & 15.20 
& \cellcolor{orange!20}96.67 & 9.83
& \cellcolor{orange!20}97.83 & 11.22
& \cellcolor{orange!20}98.55 & 12.53
& \cellcolor{orange!20}94.50 & 11.27
& \cellcolor{orange!20}99.00 & 12.04\\
\midrule
& \multicolumn{2}{c|}{MDOCVRPBL}
& \multicolumn{2}{c|}{MDOCVRPMBL}
& \multicolumn{2}{c|}{MDCVRPBLTW}
& \multicolumn{2}{c|}{MDCVRPMBLTW}
& \multicolumn{2}{c|}{MDOCVRPBLTW}
& \multicolumn{2}{c}{MDOCVRPMBLTW} \\

LLM
& 42.45 & 9.95
& 60.00 & 8.12
& 31.02 & 9.78
& 61.95 & 12.39
& 14.03 & 13.53
& 30.56 & 13.13 \\

Self-Guided
& \cellcolor{orange!20}100.00 & 10.78
& \cellcolor{orange!20}99.52 & 10.23
& \cellcolor{orange!20}98.61 & 12.43
& \cellcolor{orange!20}100.00 & 11.60
& \cellcolor{orange!20}99.10 & 15.28
& \cellcolor{orange!20}98.46 & 14.52 \\
\bottomrule
\end{tabular}
\end{threeparttable}
\label{Tab:fault_48}
}

\end{table*}
\subsection{Insight Generation}
The framework then proceeds to the insight generation stage, where the LLM derives structured insights for error analysis. A detailed example of the prompt used at this stage is provided in Appendix~\ref{sec:prompt_Insight Generation}. Based on the previously generated instance analysis summary, the LLM infers which parts of the instance are likely to contain errors. More specifically, it reasons about the potential locations of formatting errors, such as missing sections or invalid values, as well as feasibility errors, such as capacity violations, and then specifies a corresponding reasoning path.

Following this reasoning path, the LLM generates a set of function-level prompts, each corresponding to a specific analysis step. These prompts clearly describe the intended functionality, required inputs, and expected outputs, and are used to guide code generation. During this process, the LLM is instructed to ensure that the generated code reports detailed diagnostic information at the point of error detection, allowing for precise localization and subsequent revision.

The generated code is then executed to perform structured instance analysis and extract concrete evidence of potential errors. If execution failures or runtime bugs occur, the resulting bug messages are fed back to the LLM, which revises the code accordingly. Once execution succeeds, the analysis outputs are collected and summarized as error insights, which form the basis for subsequent reporting and decision-making.

Overall, in this stage, the LLM guides code generation to convert its reasoning into executable procedures.
This design offers two key advantages: (i) complex constraint calculations are offloaded to code execution, addressing the limitations of LLMs in precise numerical reasoning; and
(ii) programmatic analysis enables both efficient and comprehensive evaluation of the instance,  substantially reducing the reasoning burden on the LLM.

\subsection{Report Generation}
In the final stage, the framework performs report generation by synthesizing the derived insights with the original instance information. Guided by the extracted error insights, the LLM systematically aligns the identified errors with the corresponding components of the input instance. 
This enables the generation of a comprehensive and structured report that clearly outlines the types, locations, and potential causes of the detected errors. Based on this report, the LLM is further instructed to revise the instance accordingly, correcting the identified errors to obtain a valid and consistent instance specification. The prompt used for this stage is provided in Appendix~\ref{sec:prompt_Report Generation}.

\section{Experiment}
To evaluate the effectiveness of the proposed self-guided framework, we compare it against LLM across 48 VRP variants. Then we conduct a comprehensive analysis along multiple dimensions, including reasoning depth, error number, instance scale, and constraint complexity, to assess the performance improvements introduced by our framework. Furthermore, we provide comparisons with several prompting strategies to further validate the advantages of our approach. Finally, we extend our evaluation to other combinatorial optimization problems and LLMs to demonstrate the generality of the proposed framework. 

To provide a more complete understanding of the framework's behavior, additional analyses are presented in the appendix, including false positive analysis on valid instances (Appendix~\ref{sec:False Positive Analysis on Valid Instances}), comparison of error detection accuracy on instance-level (Appendix~\ref{sec:Comparison of Instance-Level Error Detection Accuracy}), and downstream solver validation on repaired instances (Appendix~\ref{sec:Downstream Solver Validation on Repaired Instances}). And details on hyperparameters and the experimental setup are provided in Appendix~\ref{sec:Hyperparameters and Device}.
\begin{table*}[ht]
\centering
\vspace{-4mm}
\caption{Impact of instance scale and fault number on EDR for LLM and the Self-Guided framework.}
\vspace{2mm}
\renewcommand\arraystretch{1.2}
\resizebox{\textwidth}{!}{
\begin{tabular}{l|cc|cc|cc|cc|cc|cc}
\toprule
Scale
& \multicolumn{6}{c|}{50} 
& \multicolumn{6}{c}{100} \\
\midrule
Fault Number 
& \multicolumn{2}{c|}{5}& \multicolumn{2}{c|}{10}& \multicolumn{2}{c|}{15}
& \multicolumn{2}{c|}{5}& \multicolumn{2}{c|}{10}& \multicolumn{2}{c}{15} \\
\midrule
EDR(\%)& LLM & Self-Guided
& LLM & Self-Guided 
& LLM & Self-Guided 
& LLM & Self-Guided
& LLM & Self-Guided 
& LLM & Self-Guided\\
\midrule

CVRP
& 97.00 & \cellcolor{orange!20}100.00
& 87.25 & \cellcolor{orange!20}99.49
& 86.82 & \cellcolor{orange!20}98.99
& 94.00 & \cellcolor{orange!20}98.00
& 89.80 & \cellcolor{orange!20}97.96
& 82.33 & \cellcolor{orange!20}97.90 \\

CVRPTW
& 79.00 & \cellcolor{orange!20}97.00
& 66.50 & \cellcolor{orange!20}95.00
& 63.67 & \cellcolor{orange!20}90.33
& 68.00 & \cellcolor{orange!20}98.00
& 64.00 & \cellcolor{orange!20}93.00
& 62.33 & \cellcolor{orange!20}87.67 \\

CVRPB
& 82.00 & \cellcolor{orange!20}94.00
& 78.50 & \cellcolor{orange!20}94.00
& 56.02 & \cellcolor{orange!20}91.73
& 76.00 & \cellcolor{orange!20}95.00
& 77.00 & \cellcolor{orange!20}92.00
& 53.33 & \cellcolor{orange!20}92.33 \\

OCVRPB
& 93.00 & \cellcolor{orange!20}95.00
& 84.00 & \cellcolor{orange!20}94.00
& 71.62 & \cellcolor{orange!20}88.85
& 85.00 & \cellcolor{orange!20}99.00
& 79.00 & \cellcolor{orange!20}89.00
& 64.33 & \cellcolor{orange!20}87.00 \\

CVRPMBL
& 70.05 & \cellcolor{orange!20}99.00
& 65.00 & \cellcolor{orange!20}91.50
& 47.52 & \cellcolor{orange!20}88.37
& 67.74 & \cellcolor{orange!20}100.00
& 60.34 & \cellcolor{orange!20}93.50
& 45.97 & \cellcolor{orange!20}87.58 \\

MDCVRPMBTW
& 70.00 & \cellcolor{orange!20}92.00
& 59.50 & \cellcolor{orange!20}90.00
& 54.67 & \cellcolor{orange!20}88.33
& 69.00 & \cellcolor{orange!20}98.00
& 56.50 & \cellcolor{orange!20}91.00
& 52.67 & \cellcolor{orange!20}87.67 \\

MDOCVRPBTW
& 66.00 & \cellcolor{orange!20}96.00
& 60.00 & \cellcolor{orange!20}94.50
& 49.67 & \cellcolor{orange!20}86.67
& 58.00 & \cellcolor{orange!20}100.00
& 55.50 & \cellcolor{orange!20}95.50
& 48.33 & \cellcolor{orange!20}94.66 \\

MDOCVRPMBLTW
& 72.59 & \cellcolor{orange!20}98.56
& 69.44 & \cellcolor{orange!20}98.46
& 44.67 & \cellcolor{orange!20}96.00
& 54.98 & \cellcolor{orange!20}97.26
& 44.46 & \cellcolor{orange!20}97.85
& 39.87 & \cellcolor{orange!20}96.46 \\

\bottomrule
\end{tabular}
}
\label{tab:number_scale}
\end{table*}

\begin{table*}[t]
\centering
\caption{Impact of Reasoning Depth on EDR for LLM and the Self-Guided Framework}
\vspace{2mm}
\renewcommand\arraystretch{1.2}
\resizebox{\textwidth}{!}{
\begin{tabular}{l|cc|cc|cc|cc|cc}
\toprule
Reasoning Depth
& \multicolumn{2}{c|}{1} 
& \multicolumn{2}{c|}{2} 
& \multicolumn{2}{c|}{3} 
& \multicolumn{2}{c|}{4} 
& \multicolumn{2}{c}{5} \\
EDR(\%)& LLM & Self-Guided
& LLM & Self-Guided 
& LLM & Self-Guided 
& LLM & Self-Guided
& LLM & Self-Guided \\
\midrule
CVRP
& 87.25 & \cellcolor{orange!20}99.49
& 94.89 & \cellcolor{orange!20}100.00
& 96.42 & \cellcolor{orange!20}100.00
& 97.45 & \cellcolor{orange!20}100.00
& 97.96 & \cellcolor{orange!20}100.00 \\

CVRPTW
& 66.50 & \cellcolor{orange!20}95.00
& 68.00 & \cellcolor{orange!20}98.00
& 69.00 & \cellcolor{orange!20}98.00
& 70.00 & \cellcolor{orange!20}99.00
& 72.50 & \cellcolor{orange!20}100.00 \\

CVRPB
& 78.50 & \cellcolor{orange!20}94.00
& 85.00 & \cellcolor{orange!20}96.00
& 90.00 & \cellcolor{orange!20}96.00
& 94.00 & \cellcolor{orange!20}99.00
& 99.00 & \cellcolor{orange!20}100.00 \\

OCVRPB
& 84.00 & \cellcolor{orange!20}94.00
& 88.50 & \cellcolor{orange!20}99.50
& 90.50 & \cellcolor{orange!20}100.00
& 93.00 & \cellcolor{orange!20}100.00
& 94.50 & \cellcolor{orange!20}100.00 \\

CVRPMBL
& 65.00 & \cellcolor{orange!20}91.50
& 67.50 & \cellcolor{orange!20}95.50
& 69.00 & \cellcolor{orange!20}100.00
& 70.00 & \cellcolor{orange!20}100.00
& 70.50 & \cellcolor{orange!20}100.00 \\

MDCVRPMBTW
& 59.50 & \cellcolor{orange!20}90.00
& 60.00 & \cellcolor{orange!20}99.00
& 60.00 & \cellcolor{orange!20}99.50
& 61.00 & \cellcolor{orange!20}100.00
& 65.50 & \cellcolor{orange!20}100.00 \\

MDOCVRPBTW
& 60.00 & \cellcolor{orange!20}94.50
& 61.50 & \cellcolor{orange!20}98.50
& 62.00 & \cellcolor{orange!20}99.00
& 62.50 & \cellcolor{orange!20}99.00
& 64.00 & \cellcolor{orange!20}100.00 \\

MDOCVRPMBLTW
& 30.56 & \cellcolor{orange!20}98.46
& 31.00 & \cellcolor{orange!20}99.54
& 34.93 & \cellcolor{orange!20}100.00
& 35.26 & \cellcolor{orange!20}100.00
& 36.24 & \cellcolor{orange!20}100.00 \\
\bottomrule
\end{tabular}
}
\label{tab:llm_iteration}
\end{table*}

\begin{table*}[ht]
\vspace{-4mm}
\centering
\caption{Comparison with Different Prompt Strategies.}
\renewcommand\arraystretch{1.2}
\resizebox{\textwidth}{!}{
\begin{tabular}{l|cc|cc|cc|cc|cc}
\toprule
& \multicolumn{2}{c|}{LLM}
& \multicolumn{2}{c|}{Self-Refine}
& \multicolumn{2}{c|}{Critic-Guided Refinement}
& \multicolumn{2}{c|}{CoT}
& \multicolumn{2}{c}{Self-Guided} \\
\midrule
& EDR(\%) & Time(m)
& EDR(\%) & Time(m)
& EDR(\%) & Time(m)
& EDR(\%) & Time(m)
& EDR(\%) & Time(m) \\
\midrule

CVRP
& 87.25 & 6.35
& 92.86 & 9.60
& 92.35 & 9.12
& 91.33 & 7.13
& \cellcolor{orange!20}99.49 & 6.39 \\

CVRPTW
& 66.50 & 7.00
& 69.00 & 11.66
& 68.50 & 13.92
& 69.00 & 8.10
& \cellcolor{orange!20}95.00 & 7.22 \\

CVRPB
& 78.50 & 8.25
& 82.00 & 11.43
& 83.50 & 10.52
& 83.00 & 7.60
& \cellcolor{orange!20}94.00 & 9.03 \\

OCVRPB
& 84.00 & 8.55
& 84.50 & 11.33
& 86.00 & 11.08
& 85.00 & 6.78
& \cellcolor{orange!20}94.00 & 8.52 \\

CVRPMBL
& 65.00 & 7.63
& 69.00 & 10.12
& 70.50 & 10.50
& 67.00 & 5.88
& \cellcolor{orange!20}91.50 & 7.93 \\

MDCVRPMBTW
& 59.50 & 14.42
& 65.50 & 12.60
& 60.00 & 14.73
& 62.00 & 9.05
& \cellcolor{orange!20}90.00 & 15.20 \\

MDOCVRPBTW
& 60.00 & 10.53
& 62.00 & 11.61
& 61.50 & 14.96
& 60.50 & 9.83
& \cellcolor{orange!20}94.50 & 11.27 \\

MDOCVRPMBLTW
& 30.56 & 10.13
& 34.89 & 13.20
& 31.88 & 16.00
& 33.18 & 9.15
& \cellcolor{orange!20}98.46 & 14.52 \\

\bottomrule
\end{tabular}
}
\label{tab:prompt_strategy}
\end{table*}

\subsection{Comparison of Error Detection Rate}
We conduct experiments on 48 VRP variants to assess error detection performance. The comparison of Error Detection Rate (EDR) is reported in Table~\ref{Tab:fault_48}. As shown by these results, directly applying LLMs exhibits limited error detection capability on many variants, especially those involving multiple constraints or complex combinations, where the EDR often falls below 70\% and in some cases even below 50\%. 

By contrast, the proposed self-guide approach consistently achieves improvements in error detection performance across all variants. In the most cases, self-guide raises the EDR to above 95\%, and frequently reaches near-perfect detection rates, demonstrating its robustness under diverse problem settings and constraint combinations. Importantly, these performance gains are obtained without introducing significant additional computational overhead. As shown by the time statistics, self-guide only incurs a marginal increase in runtime compared with direct LLM inference, while delivering dramatically higher error detection accuracy.

\subsection{Performance Analysis}

\textbf{Impact of Error Number.} 
Table~\ref{tab:number_scale} shows that increasing the number of injected errors significantly degrades the error detection performance of direct LLM inference, especially for instances with complex constraints such as CVRPMBL and MDOCVRPMBLTW. In contrast, the self-guided framework remains robust to error number variations, consistently achieving substantially higher EDR and often maintaining over 90\% detection accuracy even when the error number reaches 15.

\textbf{Impact of Instance Scale.} The impact of instance scale is also evident from Table~\ref{tab:number_scale}. When moving from 50-node to 100-node instances, the error detection performance of direct LLM inference generally deteriorates, especially under larger error numbers. This performance drop highlights the difficulty of handling larger problem instances with more complex interactions using pure LLM-based reasoning. In contrast, the self-guided framework exhibits much better scalability. While instance scale increases the overall difficulty of the task, self-guided framework maintains consistently high EDR across most variants, with only moderate performance degradation in the most challenging settings. This demonstrates that the proposed framework scales more effectively with instance size, enabling robust error detection even on larger and more complex VRP instances.

\textbf{Effect of Reasoning Depth.} Table~\ref{tab:llm_iteration} reports the impact of reasoning depth on error detection performance for both direct LLM inference and the proposed self-guided framework. As the reasoning depth increases, direct LLM inference shows only limited and gradual improvements, and fails to consistently achieve perfect error detection across different VRP variants. In contrast, the self-guided framework exhibits clear and stable convergence behavior. Under our framework, a reasoning depth of five is sufficient to guarantee a 100\% error detection rate across all evaluated VRP variants.

A detailed analysis of the \textbf{impact of constraints type count} is provided in Appendix~\ref{sec:Impact of Constraints}.
\subsection{Comparison with Prompt Strategies}
To demonstrate the advantage of self-guided framework, we compare it with self-refine~\citep{madaan2023self}, critic-guided refinement~\citep{bai2022constitutional} and COT~\citep{wei2022chain}. All methods are evaluated under the same experimental settings to ensure a fair comparison.

As shown in Table~\ref{tab:prompt_strategy}, while Self-Refine, critic-guided refinement, and CoT consistently improve upon direct LLM inference, their gains remain limited across complex VRP variants. In contrast, the self-guided framework achieves substantially higher EDR across all tested instances, particularly on more challenging variants. Moreover, this performance improvement is obtained with only modestly increased runtime, demonstrating that the self-guided framework offers a more effective and robust solution for error detection.

\subsection{Evaluation on Other Combinatorial Optimization Problems}
\begin{table}[t]
\centering
\caption{EDR and Runtime on Bin Packing Problems.}
\renewcommand\arraystretch{1.2}
\resizebox{0.9\linewidth}{!}{
\begin{tabular}{l|cc|cc}
\toprule
 & \multicolumn{2}{c|}{LLM}
   & \multicolumn{2}{c}{Code} \\
\midrule
 & EDR(\%) & Time(m)
 & EDR(\%) & Time(m) \\
\midrule
1D BPP
 & 94.50 & 5.88
 & \cellcolor{blue!20}97.50 & 6.00\\

2D BPP
 & 90.00 & 6.97
 & \cellcolor{blue!20}94.50 & 7.35\\
\bottomrule
\end{tabular}
}
\label{tab:bpp_results}
\end{table}

We additionally apply our framework to the Bin Packing Problem (BPP). BPP is a classical combinatorial optimization problem that aims to pack a set of items into the minimum number of bins subject to capacity constraints. In the one-dimensional setting (1D BPP), each item is characterized by a single size, whereas in the two-dimensional setting (2D BPP), items have both width and height, introducing additional geometric constraints and increasing problem complexity.

As shown in Table~\ref{tab:bpp_results}, the proposed self-guided framework consistently outperforms direct LLM inference on both 1D and 2D BPP. While the LLM already achieves relatively high error detection rates, our framework further improves performance, reaching 97.50\% on 1D BPP and 94.50\% on 2D BPP. These improvements are obtained with only a marginal increase in runtime, demonstrating that the self-guided framework can effectively extend to other combinatorial optimization problems.

\subsection{Application to Other LLMs}
\begin{table}[ht]
\centering
\caption{Comparison across different LLMs with and without the Self-Guided framework on EDR (\%).}
\renewcommand\arraystretch{1.2}
\resizebox{0.48\textwidth}{!}{
\begin{tabular}{l|cc|cc|cc}
\toprule
& \multicolumn{2}{c|}{GPT-4.1} 
& \multicolumn{2}{c|}{GPT-4o} 
& \multicolumn{2}{c}{Claude-4} \\
\midrule
& LLM & Self-Guided 
& LLM & Self-Guided 
& LLM & Self-Guided \\
\midrule
CVRP
& 87.25 & \cellcolor{blue!20}99.49
& 85.71 & \cellcolor{blue!20}98.99
& 91.84 & \cellcolor{blue!20}100.00 \\

CVRPTW
& 66.50 & \cellcolor{blue!20}95.00
& 60.00 & \cellcolor{blue!20}93.50
& 68.50 & \cellcolor{blue!20}96.00 \\

CVRPB
& 78.50 & \cellcolor{blue!20}94.00
& 75.00 & \cellcolor{blue!20}93.00
& 80.00 & \cellcolor{blue!20}95.50 \\

OCVRPB
& 84.00 & \cellcolor{blue!20}94.00
& 85.50 & \cellcolor{blue!20}95.00
& 86.50 & \cellcolor{blue!20}95.50 \\

CVRPMBL
& 65.00 & \cellcolor{blue!20}91.50
& 63.00 & \cellcolor{blue!20}90.00
& 66.50 & \cellcolor{blue!20}92.00 \\

MDCVRPMBTW
& 59.50 & \cellcolor{blue!20}90.00
& 55.00 & \cellcolor{blue!20}88.50
& 61.00 & \cellcolor{blue!20}92.00 \\

MDOCVRPBTW
& 60.00 & \cellcolor{blue!20}94.50
& 58.50 & \cellcolor{blue!20}93.50
& 61.50 & \cellcolor{blue!20}95.00 \\

MDOCVRPMBLTW
& 30.56 & \cellcolor{blue!20}98.46
& 26.85 & \cellcolor{blue!20}94.45
& 34.07 & \cellcolor{blue!20}99.54 \\
\bottomrule
\end{tabular}
}
\vspace{-2mm}
\label{tab:other_llms}
\end{table}

We further evaluate the proposed self-guided framework on GPT-4o~\citep{openai2024gpt4o} and Claude Opus 4~\citep{anthropic2024claude4} to assess its generality across LLM backends. As shown in Table~\ref{tab:other_llms}, direct inference degrades substantially on complex VRP variants across GPT-4.1, GPT-4o, and Claude Opus 4, especially under multiple coupled constraints and long-horizon reasoning, suggesting that this limitation is not model-specific.

In contrast, the proposed self-guided framework consistently yields substantial improvements in EDR across all models and problem variants. Notably, even on the most challenging instances, self-guided inference raises the EDR to above 90\% in most cases, demonstrating stable gains across different LLM families. These results suggest that the proposed framework is largely model-agnostic and can effectively enhance error detection reliability regardless of the underlying LLM backbone.

\section{Conclusion}
\label{sec:conclusion}
In this work, we investigate error detection in vehicle routing problem instances and identifies fundamental limitations of large language models (LLMs) when confronted with instances or errors of higher complexity. 
To address these challenges, we introduce a self-guided reasoning framework that enables LLMs to generate reasoning paths to guide code generation and produce additional insights, thus enhancing their own reasoning capabilities.
Extensive experiments across diverse problem settings and LLM backbones demonstrate that the proposed approach consistently and substantially improves error detection performance, achieving near-perfect accuracy and underscoring its effectiveness and generality. 

Despite its effectiveness, the proposed self-guided framework has certain limitations. While we evaluate the framework on a wide range of COPs and LLM backbones, further validation on additional problem domains and real-world noisy datasets would be valuable to fully assess its generalization. Moreover, some errors in problem instances may interfere with the detection of other errors; for example, missing or malformed sections can obscure subsequent error detection. Although our framework alleviates this issue through iterative reasoning, such cascading effects cannot be entirely eliminated in the initial reasoning round.

\clearpage
\bibliography{arxiv}

\clearpage

\clearpage
\appendix

\section{Vehicle Routing Problems}
\label{sec:Vehicle_Routing_Problems}
Vehicle Routing Problems (VRPs) are a fundamental class of combinatorial optimization problems (COPs) that aim to design a set of vehicle routes to serve a collection of customers while minimizing a given objective, typically the total travel distance or cost. In a standard VRP setting, all routes start and end at a depot, each customer must be visited exactly once, and various operational constraints must be satisfied. Due to the rich combination of complex constraints and diverse real-world application scenarios, VRPs are commonly used as a representative benchmark for COPs.

Several constraints are commonly studied in VRPs, including capacity, open route, and multi-depot. The details of these constraints are summarized in Table~\ref{tab:vrp_constraints}, and the relationships between constraints and VRP variants are illustrated in Table~\ref{tab:vrp_variants_constraints}.
\begin{table}[h]
\centering
\caption{Summary of VRP constraints considered in this work.}
\vspace{2mm}
\renewcommand\arraystretch{1.2}
\resizebox{\linewidth}{!}{
\begin{tabular}{c|p{\linewidth}}
\toprule
\textbf{Constraint} & \textbf{Description} \\
\midrule
C  & Capacity constraint, which limits the total demand served by each vehicle so that it does not exceed its carrying capacity. \\
\midrule
O  & Open route constraint, where vehicles are not required to return to the depot after completing their routes. \\
\midrule
B  & Backhaul constraint, where customers are partitioned into linehaul and backhaul sets, and vehicles are required to serve all linehaul customers before visiting any backhaul customers on each route. \\
\midrule
MB & Mixed backhaul constraint, which relaxes strict service ordering and allows limited interleaving of linehaul and backhaul customers. \\
\midrule
L  & Distance limit constraint, which restricts the maximum travel distance of each route. \\
\midrule
TW & Time window constraint, requiring each customer to be served within a specified time interval. \\
\midrule
MD & Multi-depot constraint, which allows vehicles to start from multiple depots rather than a single centralized one. \\
\bottomrule
\end{tabular}
}
\label{tab:vrp_constraints}
\end{table}

\begin{table*}[ht]
\centering
\caption{VRP variants and their associated constraints.}
\vspace{2mm}
\renewcommand\arraystretch{1.2}
\resizebox{\textwidth}{!}{
\begin{tabular}{l|ccccccc}
\toprule
\textbf{VRP Variant} 
& \begin{tabular}{c} C \\ Capacity \end{tabular}
& \begin{tabular}{c} O \\ Open \end{tabular}
& \begin{tabular}{c} B \\ Backhaul \end{tabular}
& \begin{tabular}{c} MB \\ Mixed Backhaul \end{tabular}
& \begin{tabular}{c} L \\ Distance Limit \end{tabular}
& \begin{tabular}{c} TW \\ Time Windows \end{tabular}
& \begin{tabular}{c} MD \\ Multi-Depot \end{tabular} \\
\midrule
CVRP            & \checkmark &  &  &  &  &  &  \\
OCVRP            & \checkmark & \checkmark &  &  &  &  &  \\
CVRPB            & \checkmark &  & \checkmark &  &  &  &  \\
CVRPL            & \checkmark &  &  &  & \checkmark &  &  \\
CVRPTW           & \checkmark &  &  &  &  & \checkmark &  \\
OCVRPTW          & \checkmark & \checkmark &  &  &  & \checkmark &  \\
OCVRPB           & \checkmark & \checkmark & \checkmark &  &  &  &  \\
OCVRPL           & \checkmark & \checkmark &  &  & \checkmark &  &  \\
CVRPBL           & \checkmark &  & \checkmark &  & \checkmark &  &  \\
CVRPBTW          & \checkmark &  & \checkmark &  &  & \checkmark &  \\
CVRPLTW          & \checkmark &  &  &  & \checkmark & \checkmark &  \\
OCVRPBL          & \checkmark & \checkmark & \checkmark &  & \checkmark &  &  \\
OCVRPBTW         & \checkmark & \checkmark & \checkmark &  &  & \checkmark &  \\
OCVRPLTW         & \checkmark & \checkmark &  &  & \checkmark & \checkmark &  \\
CVRPBLTW         & \checkmark &  & \checkmark &  & \checkmark & \checkmark &  \\
OCVRPBLTW        & \checkmark & \checkmark & \checkmark &  & \checkmark & \checkmark &  \\
CVRPMB           & \checkmark &  & \checkmark & \checkmark &  &  &  \\
OCVRPMB          & \checkmark & \checkmark & \checkmark & \checkmark &  &  &  \\
CVRPMBL          & \checkmark &  & \checkmark & \checkmark & \checkmark &  &  \\
CVRPMBTW         & \checkmark &  & \checkmark & \checkmark &  & \checkmark &  \\
OCVRPMBL         & \checkmark & \checkmark & \checkmark & \checkmark & \checkmark &  &  \\
OCVRPMBTW        & \checkmark & \checkmark & \checkmark & \checkmark &  & \checkmark &  \\
CVRPMBLTW        & \checkmark &  & \checkmark & \checkmark & \checkmark & \checkmark &  \\
OCVRPMBLTW       & \checkmark & \checkmark & \checkmark & \checkmark & \checkmark & \checkmark &  \\
MDCVRP          & \checkmark &  &  &  &  &  & \checkmark \\
MDOCVRP          & \checkmark & \checkmark &  &  &  &  & \checkmark \\
MDCVRPB          & \checkmark &  & \checkmark &  &  &  & \checkmark \\
MDCVRPL          & \checkmark &  &  &  & \checkmark &  & \checkmark \\
MDCVRPTW         & \checkmark &  &  &  &  & \checkmark & \checkmark \\
MDOCVRPTW        & \checkmark & \checkmark &  &  &  & \checkmark & \checkmark \\
MDOCVRPB         & \checkmark & \checkmark & \checkmark &  &  &  & \checkmark \\
MDOCVRPL         & \checkmark & \checkmark &  &  & \checkmark &  & \checkmark \\
MDCVRPBL         & \checkmark &  & \checkmark &  & \checkmark &  & \checkmark \\
MDCVRPBTW        & \checkmark &  & \checkmark &  &  & \checkmark & \checkmark \\
MDCVRPLTW        & \checkmark &  &  &  & \checkmark & \checkmark & \checkmark \\
MDOCVRPBL        & \checkmark & \checkmark & \checkmark &  & \checkmark &  & \checkmark \\
MDOCVRPBTW       & \checkmark & \checkmark & \checkmark &  &  & \checkmark & \checkmark \\
MDOCVRPLTW       & \checkmark & \checkmark &  &  & \checkmark & \checkmark & \checkmark \\
MDCVRPBLTW       & \checkmark &  & \checkmark &  & \checkmark & \checkmark & \checkmark \\
MDOCVRPBLTW      & \checkmark & \checkmark & \checkmark &  & \checkmark & \checkmark & \checkmark \\
MDCVRPMB         & \checkmark &  & \checkmark & \checkmark &  &  & \checkmark \\
MDOCVRPMB        & \checkmark & \checkmark & \checkmark & \checkmark &  &  & \checkmark \\
MDCVRPMBL        & \checkmark &  & \checkmark & \checkmark & \checkmark &  & \checkmark \\
MDCVRPMBTW       & \checkmark &  & \checkmark & \checkmark &  & \checkmark & \checkmark \\
MDOCVRPMBL       & \checkmark & \checkmark & \checkmark & \checkmark & \checkmark &  & \checkmark \\
MDOCVRPMBTW      & \checkmark & \checkmark & \checkmark & \checkmark &  & \checkmark & \checkmark \\
MDCVRPMBLTW      & \checkmark &  & \checkmark & \checkmark & \checkmark & \checkmark & \checkmark \\
MDOCVRPMBLTW     & \checkmark & \checkmark & \checkmark & \checkmark & \checkmark & \checkmark & \checkmark \\
\bottomrule
\end{tabular}
}
\label{tab:vrp_variants_constraints}
\end{table*}

\section{Dataset}
\label{sec:dataset}
\subsection{Assumption}
We construct our error detection dataset based on benchmark instances from the Route Finder dataset~\citep{berto2025routefinder}. To evaluate the capability of LLMs in identifying instance-level errors, we systematically inject synthetic errors into valid VRP instances. To ensure that each faulty instance admits a unique and unambiguous interpretation, we impose several assumptions during dataset construction, as described below.

\begin{enumerate}
    \item[a)] \textbf{Capacity and Distance Limit Correctness.} 
    Capacity values and distance limits are assumed to be correct and are not subject to corruption. This assumption enables us to disambiguate between potential error sources, such as distinguishing demand-related errors from capacity violations, and separating node coordinate errors from distance limit misconfigurations.

    \item[b)] \textbf{Single Error Per Node.} 
    At most one error is injected per node. For example, if a node is assigned missing or invalid coordinates, no additional errors are introduced for that node. This constraint prevents overlapping errors that could lead to ambiguous error attribution.

    \item[c)] \textbf{Depot Error Exclusion.} 
    When distance limit or time window errors are present, no errors are injected at the depot. This restriction is imposed because depot errors would interfere with the computation of these constraints.
\end{enumerate}

These assumptions enable controlled error injection while avoiding cascading or interacting errors, facilitating a reliable and interpretable evaluation of error detection performance.
\subsection{Format Errors}
\begin{enumerate}
    \item[(i)] \textbf{Value Error.}
    Value errors refer to numerical values that violate basic validity constraints, such as identifiers being out of range or numerical fields taking negative values where only non-negative values are permitted. For example, a customer ID exceeding the defined index range or a negative demand value constitutes a value error in the CVRP dataset.

    \item[(ii)] \textbf{Value Missing.}
    Value missing errors occur when required numerical fields are absent in an instance. Typical examples include missing demand values, service times, or coordinates for a customer node.

    \item[(iii)] \textbf{Section Missing.}
    Section missing errors arise when the name of a required section in the instance description is absent, even though the corresponding data entries may still be present.

    \item[(vi)] \textbf{Duplicate Node.}
    Duplicate node errors occur when the same node appears multiple times in an instance.
\end{enumerate}

\subsection{Feasibility Errors}
\begin{enumerate}
    \item[(i)] \textbf{Capacity (C).} 
    This constraint is associated with exceed capacity errors, which occur when the total demand assigned to a vehicle exceeds its capacity limit. For LLMs, detecting such errors mainly requires a straightforward comparison between the demand and the capacity value. As a result, this error type is relatively easy to identify, and LLMs typically achieve high detection accuracy.

    \item[(ii)] \textbf{Backhaul (B) and Mixed Backhaul (MB).} 
    For backhaul and mixed backhaul constraints, customer demands may take positive (linehaul) or negative (backhaul) values, and feasibility requires that the absolute value of the accumulated demand along a route does not exceed the vehicle capacity. Compared with the standard capacity constraint, error detection under these settings involves an additional absolute-value operation, which slightly increases the reasoning difficulty and results in lower detection accuracy.

    \item[(iii)] \textbf{Distance Limit (L).} 
    For the distance limit constraint, we compute the maximum round-trip distance between the depot and all customer nodes, and subtract a small margin (0.1) to obtain the distance limit value. We then check whether the round-trip distance of each node to the depot falls within this limit and record related violations.
    \clearpage
    \item[(iv)] \textbf{Open Route (O).}
    Under the open route constraint, vehicles are not required to return to the depot after completing their routes. Accordingly, the error injection for the distance limit constraint is modified to compute the maximum one-way distance from the depot to all customers and subtract a small margin (0.1) to define the new distance limit value and record violations.

    \item[(v)] \textbf{Time Window (TW).}
    For the time window constraint, we randomly select a node that has not been assigned any error and compute its travel distance from the depot. The left bound of the time window is set to 0.7 times this distance, and the right bound is set to 0.9 times the same distance, which is designed to make the time window infeasible for that node.

    \item[(vi)] \textbf{Multi-Depot (MD).} 
    Under the multi-depot setting, distance-related constraints are evaluated with respect to the nearest depot. Specifically, the distance limit is computed based on the maximum round-trip distance between each node and its closest depot, rather than a single fixed depot. Similarly, when injecting time window errors, the travel distance used to construct the time window bounds is calculated with respect to the nearest depot to the selected node.
\end{enumerate}

\section{More Experiment Details}
\subsection{Hyperparameters and Device.} 
\label{sec:Hyperparameters and Device}
For error detection evaluation, we adopt benchmark instances from the RouteFinder dataset~\citep{berto2025routefinder}. For each dataset consisting of 20 instances with 50 nodes, we inject up to 200 errors to construct faulty datasets. For certain datasets (e.g., CVRP), the number of injected errors is correspondingly lower due to structural constraints. All experiments are conducted using GPT-4.1~\citep{openai2025gpt41} as the underlying LLM for error detection on a system equipped with an AMD EPYC 7702P CPU. We will make our code and data publicly available after publication. The reported data are averaged.
\subsection{False Positive Analysis on Valid Instances}
\label{sec:False Positive Analysis on Valid Instances}
\begin{table}[H]
\centering
\caption{Number of faults detected by LLM on correct VRP instances.}
\vspace{2mm}
\renewcommand\arraystretch{1.2}
\resizebox{1.0\linewidth}{!}{
\begin{tabular}{l|ccccccc}
\toprule
 & CVRP & OCVRP & MDCVRP & CVRPB & CVRPMB & CVRPTW & CVRPL \\
\midrule
LLM     & 2 & 1 & 0 & 11 & 13 & 4 & 0 \\
Self-Guided 
        & \cellcolor{blue!20}0 
        & \cellcolor{blue!20}0 
        & 0 
        & \cellcolor{blue!20}0  
        & \cellcolor{blue!20}0  
        & \cellcolor{blue!20}0  
        & 0 \\
\bottomrule
\end{tabular}
}
\label{tab:right}
\end{table}

We further evaluate the error detection behavior on correct VRP instances. As shown in Table~\ref{tab:right}, direct LLM inference may incorrectly report non-existent errors even when the input instance is valid, which results in false positives across several VRP variants. In contrast, the proposed self-guided framework does not exhibit this issue: when no errors are present in the instance, it consistently reports the instance as correct. This demonstrates that self-guided reasoning effectively reduces false alarms and improves the reliability of error detection.
\subsection{Comparison of Instance-Level Error Detection Accuracy}
\label{sec:Comparison of Instance-Level Error Detection Accuracy}
\begin{table}[h]
\centering
\caption{Percentage of instances where all faults are correctly detected.}
\vspace{2mm}
\renewcommand\arraystretch{1.2}
\resizebox{1.0\linewidth}{!}{
\begin{tabular}{l|ccccccc}
\toprule
(\%) & CVRP & OCVRP & MDCVRP & CVRPB & CVRPMB & CVRPTW & CVRPL \\
\midrule
LLM     & 65 & 85 & 85 & 55 & 75 & 10 & 5 \\
Self-Guided & \cellcolor{blue!20}95 & \cellcolor{blue!20}95 & \cellcolor{blue!20}95 & \cellcolor{blue!20}85 & \cellcolor{blue!20}90 & \cellcolor{blue!20}90 & \cellcolor{blue!20}100 \\
\bottomrule
\end{tabular}
}
\label{tab:all_faults_correct}
\end{table}

\begin{table*}[t]
\centering
\caption{Downstream solver performance on repaired VRP instances.}
\vspace{2mm}
\renewcommand\arraystretch{1.2}
\resizebox{1.0\linewidth}{!}{
\begin{tabular}{l|cccccccc}
\toprule
AFL& CVRP & CVRPTW & CVRPB & OCVRPB & CVRPMBL & MDCVRPMBTW & MDOCVRPBTW & MDOCVRPMBLTW \\
\midrule
Obj.     & 14.19 & 14.59 & 14.87 & 9.30 & 10.14 & 12.83 & 8.71 & 7.91 \\
Time (s) & 10.51 & 11.01 & 10.94 & 11.42 & 12.63 & 11.91 & 12.14 & 12.72 \\
\bottomrule
\end{tabular}
}
\label{tab:solver_repaired}
\end{table*}

Table~\ref{tab:all_faults_correct} reports the instance-level error detection accuracy, measured as the percentage of instances for which all errors are correctly identified. The results show that direct LLM inference struggles to achieve reliable instance-level correctness, especially for variants with complex constraints such as CVRPTW and CVRPL, where the accuracy drops to 10\% and 5\%, respectively. In contrast, the proposed self-guided framework consistently achieves higher instance-level accuracy across all VRP variants, with performance remaining above 85\% and reaching 100\% on CVRPL. These results demonstrate that self-guided reasoning substantially improves the reliability of error detection at the instance level, particularly in settings with complex constraints.
\subsection{Downstream Solver Validation on Repaired Instances}
\label{sec:Downstream Solver Validation on Repaired Instances}
We apply the instances revised by our framework to AFL~\citep{zhang2025agentic}, an LLM-based solver that takes problem instances as input and is capable of handling a wide range of VRP constraints. As shown in Table~\ref{tab:solver_repaired}, these revised instances can be directly used by AFL without any additional modification.
\subsection{Impact of Constraint Type Counts.} 
\begin{table}[H]
\centering
\caption{EDR under different counts of constraint types.}
\vspace{2mm}
\renewcommand\arraystretch{1.2}
\resizebox{\linewidth}{!}{
\begin{tabular}{c|cccccc}
\toprule
EDR(\%) & 1 & 2 & 3 & 4 & 5 & 6 \\
\midrule
LLM
& \cellcolor{blue!25}87.27
& \cellcolor{blue!20}79.42
& \cellcolor{blue!16}72.42
& \cellcolor{blue!12}57.42
& \cellcolor{blue!8}53.97
& \cellcolor{blue!5}22.30 \\

Code
& \cellcolor{blue!55}99.49
& \cellcolor{blue!50}96.90
& \cellcolor{blue!45}96.33
& \cellcolor{blue!40}95.12
& \cellcolor{blue!35}97.96
& \cellcolor{blue!30}98.78 \\
\bottomrule
\end{tabular}}
\label{tab:constraints}
\end{table}
\label{sec:Impact of Constraints}
Table~\ref{tab:constraints} shows that as the number of constraint types increases, the error detection rate of direct LLM inference degrades rapidly. In contrast, the self-guided framework remains highly robust to increasing constraint combination complexity, and consistently achieves over 95\% error detection across all settings. These results demonstrate that the proposed framework effectively mitigates the negative impact of complex constraint combination on error detection.

\subsection{Evaluation on the Hard-Mode Dataset.}
\begin{table*}[t]
\centering
\caption{Error Detection Rate (EDR, \%) on the more challenging hard-mode dataset, where multiple errors may occur within the same node.}
\label{tab:hard_mode}
\begin{tabular}{lccccccc}
\toprule
\textbf{Method} & CVRP & OCVRP & MDCVRP &
CVRPB & CVRPMB & CVRPTW & CVRPL \\
\midrule
LLM & 80.50 & 83.00 & 90.00 & 71.50 & 69.00 & 60.50 & 58.00 \\
Self-guided & \textbf{97.50} & \textbf{95.50} & \textbf{96.00} &
\textbf{89.50} & \textbf{91.00} & \textbf{89.00} & \textbf{87.00} \\
\bottomrule
\end{tabular}
\end{table*}
The controlled error-injection setting used in our main experiments (e.g., at most one error per node with fixed injection parameters) enables attribution-clean evaluation and avoids ambiguities caused by cascading errors. In real-world operational data, however, multiple faults may interact and produce more complex failure patterns. To better reflect such scenarios, we additionally construct a more challenging \emph{hard-mode} dataset, where multiple errors are allowed to occur within the same node. As shown in Table~\ref{tab:hard_mode}, the self-guided framework remains consistently more robust than direct LLM diagnosis across all evaluated VRP variants, with particularly large improvements on variants involving more complex constraints.

\subsection{Error-Type-Specific False Positive Analysis.}
\begin{table*}[t]
\centering
\caption{False-positive analysis on valid instances. Each entry denotes the number of valid instances incorrectly identified as containing the corresponding error type. ``--'' indicates that the error type is not applicable to the corresponding VRP variant.}
\label{tab:false_positive}
\begin{tabular}{lccccccc}
\toprule
\textbf{Error Type / Method} &
CVRP & OCVRP & MDCVRP &
CVRPB & CVRPMB & CVRPTW & CVRPL \\
\midrule

Format (LLM)
& 0 & 0 & -- & -- & 1 & 0 & 0 \\
Format (Self-guided)
& 0 & 0 & -- & -- & \textbf{0} & 0 & 0 \\

\midrule
Capacity (LLM)
& 2 & 1 & -- & -- & 0 & 1 & 0 \\
Capacity (Self-guided)
& \textbf{0} & \textbf{0} & -- & -- & 0 & \textbf{0} & 0 \\

\midrule
Backhaul (LLM)
& -- & -- & 11 & 12 & -- & -- & -- \\
Backhaul (Self-guided)
& -- & -- & \textbf{0} & \textbf{0} & -- & -- & -- \\

\midrule
Distance Limit (LLM)
& -- & -- & -- & -- & -- & -- & 0 \\
Distance Limit (Self-guided)
& -- & -- & -- & -- & -- & -- & 0 \\

\midrule
Time Window (LLM)
& -- & -- & -- & -- & -- & 3 & -- \\
Time Window (Self-guided)
& -- & -- & -- & -- & -- & \textbf{0} & -- \\

\bottomrule
\end{tabular}
\end{table*}
While EDR summarizes the overall error-detection performance and incorporates localization accuracy, it does not fully characterize diagnostic reliability. In particular, a practical diagnostic system should avoid incorrectly flagging valid instances as erroneous. We therefore further evaluate error-type-specific false positives on valid instances. Table~\ref{tab:false_positive} reports the number of false-positive detections for each constraint type. The self-guided framework substantially suppresses spurious diagnoses, eliminating all observed false positives in these experiments. Together with the downstream evaluation of repaired instances in Appendix~\ref{sec:Downstream Solver Validation on Repaired Instances}, these results provide a more complete assessment of the reliability of the proposed framework.

\section{Example Prompts}
\label{sec:prompt}
\subsection{Instance Analysis}
\label{sec:prompt_Instance Analysis}
\textbf{Prompt 1.1 Instance Analysis}
\begin{framed}
\small
\noindent
We need to solve a VRP instance. I will provide you with the instance. Please analyze it carefully.

\medskip
\noindent
\textbf{Step 1.} Identify the constraints of the instance and list them clearly in numbered format 1), 2), 3), \ldots{} within [ ].  
For each constraint, write both the abbreviation (if any) and a short explanation  
(e.g., \emph{Capacity (C): vehicles have limited capacity}).  
Do not include instance-specific details such as the exact number of nodes, vehicles, or capacity values.

\medskip
\noindent
\textbf{Step 2.} Write the standard problem type abbreviation  
(e.g., TSP, CVRP, CVRPL, VRPTW, OVRP, MDVRP), enclosed in quotation marks `` ''.

\medskip
\noindent
\textbf{Step 3.} List the sections that a correct and feasible instance should contain  
(e.g., \texttt{DIMENSION}, \texttt{CAPACITY}, if present).

\medskip
\noindent
\textbf{Output format:}
\begin{enumerate}
    \item \texttt{[constraints]}
    \item "specific name"
    \item \texttt{[sections]}
\end{enumerate}
\end{framed}

\noindent \textbf{Output 1.1 Instance Analysis}
\begin{framed}
\small
\noindent
\medskip
\noindent
\textbf{1.}
\texttt{[}
\begin{enumerate}[label=\arabic*)]
  \item Capacity (C): each vehicle has a limited carrying capacity that cannot be exceeded.
  \item Visit constraint (V): each customer must be visited exactly once.
  \item Depot constraint (D): all routes must start and end at the depot.
\end{enumerate}
\texttt{]}

\medskip
\noindent
\textbf{2.}
\texttt{"CVRP"}

\medskip
\noindent
\textbf{3.}
\texttt{[DIMENSION, CAPACITY, NODE\_COORD\_SECTION, DEMAND\_SECTION, DEPOT\_SECTION]}
\end{framed}

\noindent \textbf{Prompt 1.2 Self-Validation}
\begin{framed}
\small
\noindent

\medskip
\noindent
You are a VRP expert. I will give you:
\begin{enumerate}[label=\arabic*)]
    \item The original \texttt{.vrp} file content
    \item LLM's answer (constraints, specific name, and required sections)
\end{enumerate}

\medskip
\noindent
Your task is to judge whether the answer is correct. Specifically, check consistency among:
\begin{itemize}
    \item Constraints implicitly or explicitly specified in the \texttt{.vrp} file  
    \item The specific problem name vs.\ the \texttt{.vrp} file
    \item The listed constraints vs.\ the specific problem name  
\end{itemize}

\medskip
\noindent
Do not include any extra elements not present in the instance.

\medskip
\noindent
If any contradiction exists, treat the \texttt{.vrp} file as the ground truth and mark the answer as incorrect.  
If everything is consistent, mark it as correct.

\medskip
\noindent
\textbf{Output format (exactly four lines):}
\begin{enumerate}
    \item \texttt{[right: True/False]}
    \item \texttt{[jud: explanation]}
\end{enumerate}

\medskip
\noindent
\textbf{Here is the VRP file:}

\noindent
\texttt{\{vrp\_text\}}

\medskip
\noindent
\textbf{Here is LLM's answer:}

\noindent
\texttt{\{problem\_detail\}}
\end{framed}
\noindent \textbf{Output 1.2 Self-Validation}

\begin{tcolorbox}[
  colback=white,
  colframe=black,
  boxrule=0.6pt,
  arc=2pt,
  left=4pt,
  right=4pt,
  top=4pt,
  bottom=4pt,
  breakable
]
\footnotesize
\raggedright
\sloppy
\small

\medskip
\noindent
1. [right: True] \\
2. [jud: The specific problem name is consistent with the VRP file. All listed constraints are explicitly or implicitly supported by the instance content, and the required sections correspond exactly to those defined in the file. No contradictions are found, and the instance is correctly identified.]
\end{tcolorbox}

\noindent \textbf{Prompt 1.3 Revise}
\begin{tcolorbox}[
  colback=white,
  colframe=black,
  boxrule=0.6pt,
  arc=2pt,
  left=4pt,
  right=4pt,
  top=4pt,
  bottom=4pt,
  breakable
]
\footnotesize
\raggedright
\sloppy
\small
\small
\noindent
We need to solve a VRP instance. I will provide you with the instance. Please analyze it carefully.

\medskip
\noindent
\textbf{Step 1.} Identify the constraints of the instance and list them clearly in numbered format 1), 2), 3), \ldots{} within \texttt{[ ]}.  
For each constraint, write both the abbreviation (if any) and a short explanation  
(e.g., \emph{Capacity (C): vehicles have limited capacity}).  
Do not include instance-specific details such as the exact number of nodes, vehicles, or capacity values.

\medskip
\noindent
\textbf{Step 2.} Write the standard problem type abbreviation  
(e.g., TSP, CVRP, CVRPL, VRPTW, OVRP, MDVRP), enclosed in quotation marks `` ''.

\medskip
\noindent
\textbf{Step 3.} List the sections that a correct and feasible instance should contain  
(e.g., \texttt{DIMENSION}, \texttt{CAPACITY}, if present).

\medskip
\noindent
\textbf{Revision instruction.}  
Below is your previous answer, followed by identified issues. Please revise your answer strictly according to the rules above and correct all inconsistencies.

\medskip
\noindent
\textbf{Here is your previous answer:}

\noindent
\texttt{\{problem\_detail\}}

\medskip
\noindent
\textbf{However, there were some issues identified:}

\noindent
\texttt{\{judgment\}}

\medskip
\noindent
\textbf{Output format:}
\begin{enumerate}
    \item \texttt{[constraints]}
    \item "specific name"
    \item \texttt{[sections]}
\end{enumerate}
\end{tcolorbox}

\noindent \textbf{Output 1.3 Revise}
\begin{framed}
\small
\noindent
\medskip
\noindent
\textbf{1.}
\texttt{[}
\begin{enumerate}[label=\arabic*)]
  \item Capacity (C): each vehicle has a limited carrying capacity that cannot be exceeded.
  \item Visit constraint (V): each customer must be visited exactly once.
  \item Depot constraint (D): all routes must start and end at the depot.
\end{enumerate}
\texttt{]}

\medskip
\noindent
\textbf{2.}
\texttt{"CVRP"}

\medskip
\noindent
\textbf{3.}
\texttt{[DIMENSION, CAPACITY, NODE\_COORD\_SECTION, DEMAND\_SECTION, DEPOT\_SECTION]}
\end{framed}

\subsection{Insight Generation}
\label{sec:prompt_Insight Generation}
\textbf{Prompt 2.1 Reasoning Path and Function Prompt Generation}
\begin{tcolorbox}[
  colback=white,
  colframe=black,
  boxrule=0.6pt,
  arc=2pt,
  left=4pt,
  right=4pt,
  top=4pt,
  bottom=4pt,
  breakable
]
\small
\noindent
You are an expert software engineer and operations research (OR) researcher.  
Your task is to design a validator program (\texttt{program.py}) in clean,
production-ready Python that checks an instance for:
(A) format validity and (B) feasibility under the given problem type and constraints,
and produces a structured report.

\medskip
\noindent
\textbf{Context}
\begin{itemize}
  \item \textbf{Problem type:} \texttt{\{specific\_name\}}
  \item \textbf{Constraints:} \texttt{\{constraints\}}
  \item \textbf{Required sections:} \texttt{\{sections\}}
\end{itemize}

\medskip
\noindent
Below is the raw \texttt{.vrp} file content (do not alter it):  
\texttt{\{vrp\_text\}}

\medskip
\noindent
\textbf{Task}

\noindent
Think step by step and design a detailed plan for the validator program.
Your design must explicitly specify:
\begin{itemize}
  \item The checks to perform (format validity and feasibility under constraints).
  \item The functions to implement, with exactly one function per check.
\end{itemize}

\medskip
\noindent
For \textbf{each function}, provide:
\begin{itemize}
  \item its purpose and role,
  \item the enforced constraints (if applicable),
  \item precise input and output definitions,
  \item its contribution to the overall validation workflow.
\end{itemize}

\medskip
\noindent
Include a \textbf{main function} that sequentially executes all validation functions.

\medskip
\noindent
\textbf{Output format (STRICT)}

\noindent
Each output line must follow exactly this structure:

\noindent
\begin{minipage}[t]{\linewidth}
\raggedright
\texttt{
(1) [function\_name: def function\_name(inputs).\\
description and purpose: what the function does.\\
enforced\_constraints: list of constraints checked (if applicable).\\
input: expected input format.]
}
\end{minipage}

\medskip

\noindent
\begin{minipage}[t]{\linewidth}
\raggedright
\texttt{
(2) [function\_name: def function\_name(inputs).\\
description and purpose: what the function does.\\
enforced\_constraints: list of constraints checked (if applicable).\\
input: expected input format.]
}
\end{minipage}

\end{tcolorbox}
\textbf{Output 2.1 Reasoning Path and Function Prompt Generation (Partly)}
\begin{tcolorbox}[
  colback=white,
  colframe=black,
  boxrule=0.6pt,
  arc=2pt,
  left=4pt,
  right=4pt,
  top=4pt,
  bottom=4pt,
  breakable
]
\small
\footnotesize
\raggedright
\sloppy

\noindent
(1) function\_name: def parse\_vrplib\_instance(vrp\_text)  

description and purpose: Reads and parses the raw VRPLIB text, extracting metadata (NAME, TYPE, DIMENSION, CAPACITY, EDGE\_WEIGHT\_TYPE) and required sections (NODE\_COORD\_SECTION, DEMAND\_SECTION, DEPOT\_SECTION), and returns a structured dictionary for validation.  

enforced\_constraints: None (data extraction only).  

input: vrp\_text (string containing the raw VRPLIB instance).

\medskip
\noindent
(2) function\_name: def check\_required\_sections(parsed\_instance)  

description and purpose: Verifies that each required header and section appears exactly once and reports missing or duplicated entries.  

enforced\_constraints: VRPLIB format specification.  

input: parsed\_instance (dictionary returned by the parser).

\medskip
\noindent
(3) function\_name: def check\_capacity\_constraint(parsed\_instance)  

description and purpose: Checks whether any customer demand exceeds the vehicle capacity and reports violating nodes.  

enforced\_constraints: Capacity constraint (C).  

input: parsed\_instance (dictionary).

\end{tcolorbox}

\noindent \textbf{Prompt 2.2 Code Generation}
\begin{tcolorbox}[
  colback=white,
  colframe=black,
  boxrule=0.6pt,
  arc=2pt,
  left=4pt,
  right=4pt,
  top=4pt,
  bottom=4pt,
  breakable
]
\footnotesize
\raggedright
\sloppy

We are working on a VRP problem instance to generate code that judges whether the instance is correct and feasible.

\textbf{Context.}
If previously generated code is provided, it should be used as reference and may be improved or extended if necessary.

\textbf{Problem Information.}

Constraints: \texttt{\{constraints\}}.  

Specific problem name: \texttt{\{specific\_name\}}.  

Sections that a correct and feasible instance should contain: \texttt{\{sections\}}.

\textbf{Important Rules.}
\begin{itemize}
  \item The raw \texttt{.vrp} file content is provided only for context.
  \item Do not hardcode any instance-specific details such as node count, vehicle number, or coordinates.
  \item All generated functions must be general-purpose and reusable.
\end{itemize}

\textbf{Task.}

Generate validation functions as specified by the task description.

If the generated function is a \emph{main function}, it must:
\begin{itemize}
  \item Accept a command-line argument \texttt{--path} using the \texttt{argparse} library.
  \item Sequentially execute all validation functions.
  \item Print only the summary \texttt{Total faults detected: [number]}.
\end{itemize}

If the generated function validates route feasibility, it must:
\begin{itemize}
  \item Check feasibility from each customer to the depot (and back if required).
  \item Respect all declared constraints.
  \item Report each infeasible case as \texttt{FAULT: [description]}.
\end{itemize}

\textbf{Output Requirement.}

Return only valid Python code enclosed in a single \texttt{python} code block.

\end{tcolorbox}

\noindent \textbf{Output 2.2 Code Generation (Partly)}
\begin{lstlisting}[language=Python, caption={Example of route feasibility check}]
import argparse
import re
import sys
from collections import OrderedDict, defaultdict

REQUIRED_FIELDS = ["NAME", "TYPE", "DIMENSION", "CAPACITY", "EDGE_WEIGHT_TYPE"]
REQUIRED_SECTIONS = ["NODE_COORD_SECTION", "DEMAND_SECTION", "DEPOT_SECTION"]


def check_required_sections(parsed_instance):
    faults = []
    missing_fields = []
    duplicated_fields = []
    missing_sections = []
    duplicated_sections = []

    meta = parsed_instance.get('_meta', {})
    header_counts = meta.get('_header_counts', {})
    section_counts = meta.get('_section_counts', {})
    for f in REQUIRED_FIELDS:
        c = header_counts.get(f, 0)
        if c == 0:
            msg = f"FAULT: Missing required field '{f}'"
            faults.append(msg)
            missing_fields.append(f)
        elif c > 1:
            msg = f"FAULT: Duplicated field '{f}': appears {c} times"
            faults.append(msg)
            duplicated_fields.append((f, c))
    section_keys = set(parsed_instance.keys())
    for s in REQUIRED_SECTIONS:
        c = section_counts.get(s, 0)
        section_exists_with_content = s in parsed_instance and bool(parsed_instance[s])
        if c == 0 or not section_exists_with_content:
            msg = f"FAULT: Missing required section '{s}'"
            faults.append(msg)
            missing_sections.append(s)
        elif c > 1:
            msg = f"FAULT: Duplicated section '{s}': appears {c} times"
            faults.append(msg)
            duplicated_sections.append((s, c))
    return {
        'missing_fields': missing_fields,
        'duplicated_fields': duplicated_fields,
        'missing_sections': missing_sections,
        'duplicated_sections': duplicated_sections,
        'faults': faults,
        'count': len(faults)
    }
\end{lstlisting}
\noindent \textbf{Prompt 2.3 Code Revision}
\begin{tcolorbox}[
  colback=white,
  colframe=black,
  boxrule=0.6pt,
  arc=2pt,
  left=4pt,
  right=4pt,
  top=4pt,
  bottom=4pt,
  breakable
]
\footnotesize
\raggedright
\sloppy

We are working on a VRP problem instance to generate code that judges whether the instance is correct and feasible.

\textbf{Context.}
If previously generated code is provided, it is shown below and should be used as reference for correction.

\medskip
\textbf{Problem Information.}
\begin{itemize}
  \item Constraints: \texttt{\{constraints\}}
  \item Specific problem name: \texttt{\{specific\_name\}}
  \item Sections that a correct and feasible instance should contain: \texttt{\{sections\}}
\end{itemize}

\medskip
\textbf{Important Rules.}
\begin{itemize}
  \item The raw \texttt{.vrp} file content is provided only for context.
  \item Do not hardcode any instance-specific details (e.g., number of nodes, vehicle count, node coordinates).
  \item All generated functions must be general-purpose and reusable for any VRP instance.
\end{itemize}

\medskip
\textbf{Instance Content (Context Only).}
\begin{verbatim}
{vrp_text}
\end{verbatim}

\medskip
\textbf{Detected Issues and Revision Guidance.}
\begin{itemize}
  \item Issues identified in the previous code: \texttt{\{error\_msg\}}
  \item Revision suggestions: \texttt{\{jud\}}
\end{itemize}

\medskip
\textbf{Task.}
Please correct the previously generated code according to the issues above, without changing the number of functions, their names, or their function signatures.

\medskip
\textbf{Output Requirement.}
Return only valid Python code enclosed in a single
\texttt{python} code block:
\begin{verbatim}
```python
...
\end{verbatim}
\end{tcolorbox}

\subsection{Report Generation}
\label{sec:prompt_Report Generation}
\noindent \textbf{Prompt 3.1 Report Generation}
\begin{tcolorbox}[
  colback=white,
  colframe=black,
  boxrule=0.6pt,
  arc=1.5pt,
  left=4pt,
  right=4pt,
  top=4pt,
  bottom=4pt,
  breakable
]
\footnotesize
\raggedright
\sloppy

You are an expert in VRP instance validation and repair.

You are given the following information:
\begin{itemize}
  \item A correct \texttt{.vrp} file used only as a format reference. Precise numeric values (e.g., distance limits) and optional sections are not treated as ground truth.
  \item A raw \texttt{.vrp} file that may contain faults.
  \item An initial list of detected faults, which may include duplicates or omissions.
\end{itemize}

Your task is to summarize all distinct faults and provide a complete repair.

\medskip
\textbf{Problem Context.}
This instance corresponds to \texttt{\{specific\_name\}}. You must verify whether all values and structures are valid under the expected specification.

\medskip
\textbf{Summary Tasks.}
Provide the following four parts:
\begin{enumerate}
  \item Detailed fault descriptions, including where the fault occurs (section, node ID, or line).
  \item Suggested corrections, one per fault.
  \item Corrected raw \texttt{.vrp} instance text with minimal necessary changes.
\end{enumerate}

\medskip
\textbf{Critical Rules.}
\begin{itemize}
  \item Reason strictly from the raw \texttt{.vrp} content.
  \item Assume capacity and distance limit values themselves are not faulty.
  \item Do not remove or rewrite large portions unnecessarily.
  \item Keep section names, structure, and indexing intact unless invalid.
\end{itemize}

\medskip
\textbf{Input Information.}

\textbf{Format Reference Instance:}
\begin{verbatim}
{right_text}
\end{verbatim}

\textbf{Initial Faults Detected:}
\begin{verbatim}
{all_fault}
\end{verbatim}

\textbf{Raw Instance with Faults:}
\begin{verbatim}
{vrp_text}
\end{verbatim}

\medskip
\textbf{Output Format.}
Return only the following three parts, each enclosed in a separate code block:
\begin{itemize}
  \item Detailed fault descriptions.
  \item Suggested corrections.
  \item Corrected raw \texttt{.vrp} instance text.
\end{itemize}

No extra commentary is allowed outside these blocks.

\end{tcolorbox}
\noindent \textbf{Output 3.1 Report Generation (Partly)}

Detailed Fault Descriptions.
\begin{tcolorbox}[
  colback=white,
  colframe=black,
  boxrule=0.6pt,
  arc=1.5pt,
  left=4pt,
  right=4pt,
  top=4pt,
  bottom=4pt,
  breakable
]
\footnotesize
\raggedright
\sloppy
\begin{itemize}
  \item Detailed fault descriptions. The node coordinate list is not explicitly preceded by the required
  \texttt{NODE\_COORD\_SECTION} header, leading to format ambiguity and
  non-compliance with the VRPLIB specification.

  \item The depot ID \texttt{1009} specified in \texttt{DEPOT\_SECTION} is outside
  the valid node index range \texttt{1..51}, does not appear in
  \texttt{NODE\_COORD\_SECTION}, and is missing from \texttt{DEMAND\_SECTION}.

  \item The demands of nodes \texttt{27}, \texttt{37}, and \texttt{47} exceed the
  declared vehicle capacity, violating the capacity constraint of CVRP.

\end{itemize}
\end{tcolorbox}

Suggested corrections.
\begin{tcolorbox}[
  colback=white,
  colframe=black,
  boxrule=0.6pt,
  arc=1.5pt,
  left=4pt,
  right=4pt,
  top=4pt,
  bottom=4pt,
  breakable
]
\footnotesize
\raggedright
\sloppy
\begin{itemize}
  \item Insert the line \texttt{NODE\_COORD\_SECTION} immediately before the node
  coordinate list to ensure unambiguous parsing.

  \item Replace the invalid depot ID \texttt{1009} with a valid node index within
  \texttt{1..DIMENSION} (e.g., node \texttt{1}), and ensure that the depot
  appears in both \texttt{NODE\_COORD\_SECTION} and \texttt{DEMAND\_SECTION}
  with zero demand.

  \item Modify the demands of nodes \texttt{27}, \texttt{37}, and \texttt{47} so that
  each individual demand does not exceed the vehicle capacity.
\end{itemize}
\end{tcolorbox}
Corrected raw .vrp instance text.
\begin{tcolorbox}[
  colback=white,
  colframe=black,
  boxrule=0.6pt,
  arc=1.5pt,
  left=4pt,
  right=4pt,
  top=4pt,
  bottom=4pt,
  breakable
]
\footnotesize
\raggedright
\sloppy
\begin{verbatim}
NAME : corrected_instance
TYPE : CVRP
DIMENSION : 51
EDGE_WEIGHT_TYPE : EUC_2D
CAPACITY : 100

NODE_COORD_SECTION
1   12.34   56.78
2   23.45   67.89
...
51  98.76   54.32

DEMAND_SECTION
1   0
2   12
...
27  40
37  35
47  30
...
51  20

DEPOT_SECTION
1
-1
EOF
\end{verbatim}
\end{tcolorbox}

\end{document}